\documentclass[letterpaper]{article} 
\usepackage[preprint]{aaai2027}
\usepackage[hyphens]{url} 
\usepackage{graphicx} 
\usepackage{natbib} 
\usepackage{caption} 
\usepackage{amsmath}
\usepackage{amssymb}
\usepackage{booktabs}
\usepackage{multirow}
\usepackage{tcolorbox}

\newcommand{\name}{\textsc{LOCUS}}

\title{LOCUS: Local Visual Cue Search for Enhancing Fine-Grained Perception in Multimodal Large Language Models}
\author{
Zhou Tao\textsuperscript{\rm 1,\rm 2}\equalcontrib,
Fang Zhang\textsuperscript{\rm 1,\rm 2}\equalcontrib,
Zewen Ding\textsuperscript{\rm 1,\rm 2},
Shida Wang\textsuperscript{\rm 1,\rm 2},
Xiaokun Sun\textsuperscript{\rm 1,\rm 2},\\
YongXiang Hua\textsuperscript{\rm 1,\rm 2},
Haoyu Cao\textsuperscript{\rm 1,\rm 2},
Linli Xu\textsuperscript{\rm 1,\rm 2}\corresponding
}

\affiliations{
\textsuperscript{\rm 1}University of Science and Technology of China\\
\textsuperscript{\rm 2}State Key Laboratory of Cognitive Intelligence
}

\begin{document}

\maketitle

\begin{abstract}
Multimodal Large Language Models (MLLMs) remain unreliable on fine-grained visual perception, even when high-resolution inputs preserve the necessary local details. We identify this limitation as \emph{visual context rot}: decisive evidence may exist in the full image, yet fail to be reliably selected and used amid redundant visual context. We propose \name{} (\textbf{LO}cal visual \textbf{CU}e \textbf{S}earch), a training framework that teaches MLLMs to \emph{internalize} local evidence search through a verifiable proxy task. During training, \name{} provides a local crop as a visual cue and optimizes the model to recover its spatial support in the full image using an IoU-based reward. The visual cue is used only during training, leaving the standard image-question inference interface unchanged. Experiments across fine-grained perception, hallucination, general understanding, and reasoning benchmarks show that \name{} improves localization-sensitive visual understanding while preserving broad capabilities. Attention analyses further indicate stronger focus on task-relevant evidence regions, suggesting that training-time visual cue search provides an effective route to internalized fine-grained evidence selection.
\end{abstract}

\section{Introduction}
\label{sec:intro}

\begin{figure}[t]
  \centering
  \includegraphics[width=\linewidth]{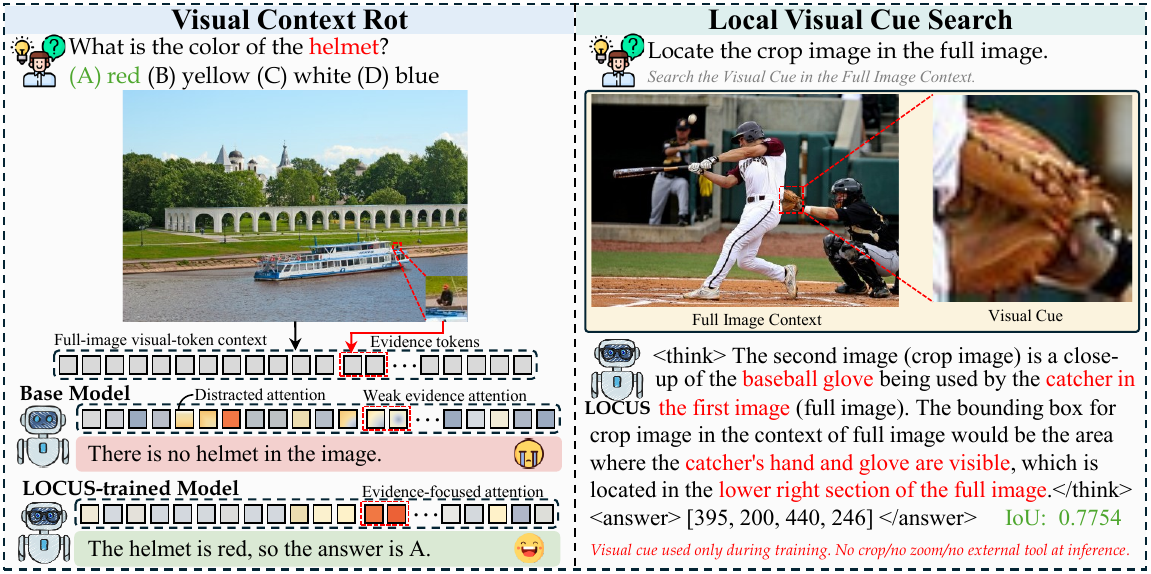}
\caption{
Motivation of \name{}. Left: fine-grained evidence occupies only a few visual tokens and is weakly attended by the base model, causing visual context rot. Right: \name{} uses training-time local visual cue search to improve evidence selection while keeping inference unchanged.
}
  \label{fig:intro}
\end{figure}

Multimodal Large Language Models (MLLMs) have achieved remarkable progress in visual understanding, demonstrating strong capabilities on a wide range of image-language tasks~\cite{bai2025qwen3vltechnicalreport,bai2025qwen25vltechnicalreport,wang2025internvl3,yue2025mimo,comanici2025gemini,team2026kimi}. Despite these advances, their performance remains fragile when correct reasoning depends on fine-grained visual evidence, such as small objects, subtle attributes, or spatially adjacent instances~\cite{wu2024v,wang2025divide,tong2024cambrian}. A natural remedy is to increase the input resolution, which preserves visual details that may otherwise be lost during downsampling. However, higher resolution also expands the visual context in which the model must search for the decisive evidence. The relevant cue often occupies only a small fraction of a long visual-token sequence, and can be diluted by surrounding objects, background semantics, and redundant observations~\cite{liu2025hide,zhang2023towards}. This reveals a fundamental gap between \emph{preserved} and \emph{accessible} visual evidence: the information required to answer a question may exist in the input, yet may not be reliably selected, retained, or exploited during standard full-image inference. We refer to this phenomenon as \emph{visual context rot}, as illustrated in Fig.~\ref{fig:intro}.

This accessibility gap is not merely a conceptual concern. Our preliminary analyses in Section~\ref{sec:preliminary_analysis} show that fine-grained accuracy improves when irrelevant visual context is suppressed while preserving the original target scale, indicating that decisive evidence is often present but obscured by surrounding context. We further observe that correctly answered samples exhibit substantially better localization of task-relevant regions than incorrectly answered ones, and that VQA accuracy increases with grounding quality. These findings suggest that fine-grained perceptual failures are closely associated with unreliable selection and use of local evidence, rather than only with the absence of visual detail. They motivate a training objective that uses localization as a verifiable proxy for strengthening the model's ability to select and exploit decisive local cues within complete images.


Existing approaches only partially address this need. Standard instruction tuning~\cite{liu2023visual,zhang2026instruction} and answer-level supervision improve final responses but provide little direct feedback on where decisive evidence lies within redundant visual context. Text-based grounding~\cite{yu2026perception,liu2025visual} introduces localization supervision, yet linguistic queries may be under-specified for small or visually similar objects. Vision-centric post-training improves general perception but does not explicitly target fine-grained evidence selection~\cite{zhan2025vision,wu2025visualjigsawposttrainingimproves}. Recent think-with-images and tool-augmented methods~\cite{zhang2025thymethinkimages,zheng2026deepeyesincentivizingthinkingimages,wang2025pixelreasonerincentivizingpixelspace,hou2026codevcodeimagesfaithful,hong2026deepeyesv2agenticmultimodalmodel} inspect localized evidence through cropping, zooming, image augmentation, or other visual manipulations. While effective, these methods retain auxiliary visual operations at inference time. Together, these limitations motivate a key question: \textit{can MLLMs be trained to internalize local evidence search, so that they can better select and use fine-grained cues during ordinary full-image inference?}

To this end, we propose \textbf{\name{}} (\textbf{LO}cal visual \textbf{CU}e \textbf{S}earch), a training framework that turns local visual cue search into a verifiable proxy task for full-image evidence use. As illustrated in Fig.~\ref{fig:intro}, during training, we crop a target region from a complete image and use it as a local visual cue, which serves as an explicit instance-level handle for the underlying evidence. Given the full image, the visual cue, and an instruction to locate the cue, the model predicts the cue's spatial support in the original image. Since the ground-truth region is known, each prediction can be directly evaluated with an IoU-based reward, allowing us to optimize the model for spatially accurate cue localization. Importantly, the visual cue is used only during training. At inference time, \name{} operates on standard image-question inputs without additional crops, zooming operations, external tools, or multi-round search.


We evaluate \name{} on a broad suite of benchmarks spanning fine-grained perception, hallucination robustness, general multimodal understanding, and mathematical and logical reasoning. The results show that \name{} consistently improves localization-sensitive fine-grained perception, with particularly strong gains on V*Bench~\cite{wu2024v} and high-resolution benchmarks, while preserving competitive performance on broad-coverage evaluations. Beyond aggregate accuracy, attention-based analyses reveal that the trained model allocates more attention to ground-truth evidence regions under standard full-image inference. Together, these results suggest that \name{} improves fine-grained perception by strengthening the model's ability to attend to and exploit decisive local evidence within full-image contexts.

Our main contributions are summarized as follows:
\begin{itemize}
    \item We characterize \emph{visual context rot} as a bottleneck in fine-grained multimodal perception, where decisive local evidence may be preserved in high-resolution inputs but not reliably selected, retained, or exploited within redundant full-image contexts. Preliminary analyses connect this phenomenon to context interference and localization quality.
    \item We propose \textbf{\name{}} (\textbf{LO}cal visual \textbf{CU}e \textbf{S}earch), a training framework that uses local visual cues to construct a verifiable proxy task for full-image evidence use. By optimizing cue localization with an IoU-based reward, \name{} strengthens local evidence search without requiring crops, zooming operations, external tools, or multi-round search at inference time.
    \item We validate \name{} across fine-grained perception, hallucination robustness, general multimodal understanding, and reasoning benchmarks. Results show consistent improvements on localization-sensitive tasks, while attention analyses indicate stronger focus on ground-truth evidence regions under standard full-image inference.
\end{itemize}

\section{Related Work}
\label{sec:related}

\paragraph{Fine-Grained Perception and Visual Search in MLLMs.}
Despite rapid progress in multimodal large language models, fine-grained visual understanding remains challenging when answers depend on small objects, subtle attributes, spatially adjacent instances, or other localized evidence that occupies only a small portion of the image~\cite{wu2024v,tong2024cambrian,wang2025divide,zhang2025mme}. This challenge has motivated \emph{Thinking-with-Images} methods that explicitly crop, zoom, search, or revisit image regions during inference~\cite{zhang2025thymethinkimages,zheng2026deepeyesincentivizingthinkingimages,wang2025pixelreasonerincentivizingpixelspace,hou2026codevcodeimagesfaithful,hong2026deepeyesv2agenticmultimodalmodel}. While effective, such methods typically rely on additional visual operations, repeated image encoding, or tool-mediated interaction at test time. Text-based grounding~\cite{yu2026perception} provides localization supervision, but linguistic queries can be under-specified for small or visually similar instances. Recent single-pass methods such as ZwZ~\cite{wei2026zooming} 
distill zoom-based inspection into ordinary inference, but rely on teacher-generated region-level QA targets. In contrast, \name{} uses the image region itself as a visual query and optimizes an automatically verifiable IoU reward for recovering its location in the original image.


\paragraph{Reinforcement Learning and Proxy Supervision for MLLMs.}
Reinforcement learning with verifiable feedback has become a prominent approach for multimodal post-training, leveraging automatically checkable task outcomes without relying on costly human preference annotation~\cite{zhang2025r1vllearningreasonmultimodal,meng2025mm}. Recent methods introduce vision-centric objectives to strengthen perception: Vision-R1~\cite{zhan2025vision} performs human-free alignment through vision-guided reinforcement learning with criterion-based visual feedback, while Visual Jigsaw~\cite{wu2025visualjigsawposttrainingimproves} trains models to recover the spatial arrangement of shuffled image patches. Related efforts optimize verifiable objectives for visual grounding~\cite{meng2025mm} and table understanding~\cite{liu2026multimodal}. ViCrit~\cite{wang2025vicritverifiablereinforcementlearning} further formulates span-level caption error identification as a verifiable proxy task for visual perception. Collectively, these studies show that controllable intermediate tasks can provide proxy supervision for transferable multimodal capabilities~\cite{zeng2026agenticjigsawinteractionlearning}. Within this broader paradigm, \name{} specifically targets instance-level spatial evidence selection through local visual cue search: given a crop and its full image, the model is rewarded by IoU for recovering the crop's spatial support. This enables the model to internalize local visual search for standard full-image inference without additional visual operations.
\section{Method}
\label{sec:method}
We present \name{}, which formulates \emph{local visual cue search}  as a proxy training task for fine-grained visual evidence discovery. During training, a local crop is used as a visual cue, and the model is required to localize its corresponding spatial support in the complete image. This formulation provides a verifiable training signal, since the predicted region can be directly evaluated by its IoU with the ground-truth box. By optimizing this task with an IoU-based reward, \name{} encourages MLLMs to internalize local visual search while retaining standard full-image inference without external visual operations.

\subsection{Preliminary Analysis}
\label{sec:preliminary_analysis}

Before detailing \name{}, we diagnose whether fine-grained failures stem from unreliable access to task-relevant local evidence under full-image context. We ask two questions: (i) does suppressing irrelevant context make decisive evidence easier to use, and (ii) are correct answers associated with better localization of the queried region?


\begin{table}[t]
\centering
\small
\setlength{\tabcolsep}{3pt}
\begin{tabular}{lccc}
\toprule
\textbf{Input View} & \textbf{Overall} & \textbf{Direct Attr.} & \textbf{Rel. Pos.} \\
\midrule
Full Image & 79.58 & 80.00 & 78.95 \\
Context-Suppressed &\textbf{85.86{$^{\uparrow 6.28}$}} &
\textbf{84.35{$^{\uparrow 4.35}$}} &
\textbf{88.16{$^{\uparrow 9.21}$}} \\
\bottomrule
\end{tabular}
\caption{Context-suppression analysis on V*Bench using Qwen2.5-VL-7B-Instruct. Non-target regions are replaced with black pixels while preserving the original target scale and image canvas.}
\label{tab:context_suppression}
\end{table}

\textbf{Suppressing irrelevant context improves fine-grained perception.}
We first test whether fine-grained failures stem from missing visual detail or from difficulty selecting relevant evidence within the full-image context. For each V*Bench sample, we preserve the ground-truth target region and replace non-target regions with black pixels, keeping the original canvas and target scale unchanged. As shown in Table~\ref{tab:context_suppression}, this context-suppressed view improves the base model from 79.58 to 85.86 overall, with gains of 4.35 on direct attributes and 9.21 on relative position. Since the target is not magnified, the improvement suggests that irrelevant context interferes with selecting and using decisive local evidence under full-image inference.

\begin{figure}[t]
  \centering
  \includegraphics[width=\linewidth]{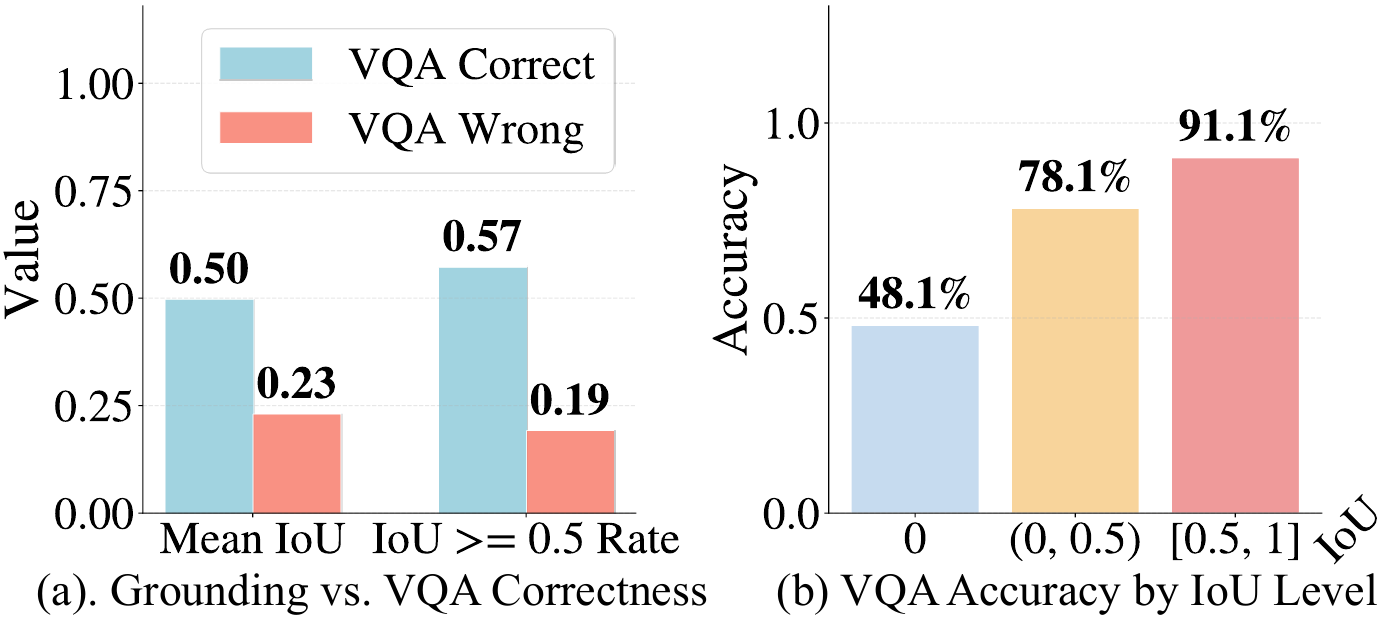} 
\caption{Grounding quality vs. VQA correctness on V*Bench direct\_attributes with Qwen2.5-VL-7B-Instruct. (a) Mean IoU and success rate (IoU $\ge$ 0.5) for correct vs. wrong answers. (b) VQA accuracy by grounding IoU.}
  \label{fig:motivation}
\end{figure}

\begin{figure*}[t]
  \centering
  \includegraphics[width=\linewidth]{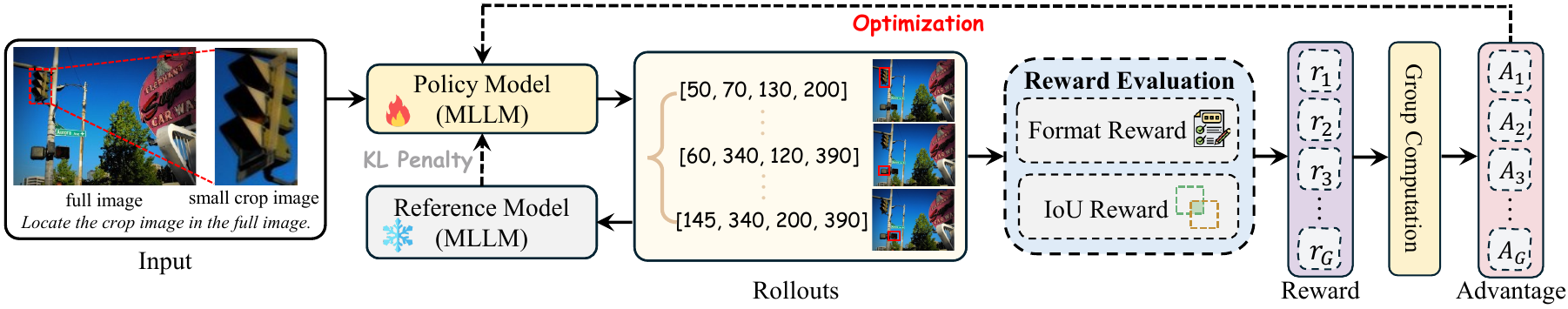} 
  \caption{Overview of \name{}. Given a full image and a localized visual exemplar cropped from it, the policy model predicts the exemplar's location in the full image. Candidate rollouts are scored by a format reward and an IoU-based localization reward, and group-relative advantages are used to optimize the policy while regularizing it against a reference model.}
  \label{fig:method}
\end{figure*}


\textbf{Grounding quality correlates with fine-grained perception.}
We next investigate the second question on the direct-attribute subset of V*Bench, where the queried object provides a well-defined grounding target. For each sample, we ask the model to localize the object referred to in the question and compare localization quality between VQA-correct and VQA-wrong samples. As shown in Fig.~\ref{fig:motivation} (Qwen3-VL results in Appendix D.1), correctly answered samples exhibit higher mean IoU and grounding success rate (IoU $\ge$ 0.5) than incorrectly answered ones. Moreover, VQA accuracy increases with grounding IoU. Samples with IoU $=0$, where the model fails to localize the queried object, achieve the lowest VQA accuracy.



Together, these analyses indicate that fine-grained perception depends on reliably accessing task-relevant local evidence within full-image context. While localization does not explain all failures, it offers a practical and verifiable proxy for strengthening evidence selection. To directly supervise local-to-global visual correspondence, we instantiate this proxy as local visual cue search: the target region itself serves as an instance-specific visual cue, and the model learns to recover its spatial support in the complete image.

\subsection{Local Visual Cue Search}
\label{sec:local_visual_cue_search}

The overall training pipeline of \name{} is summarized in Fig.~\ref{fig:method}. Given a complete image and a local visual cue cropped from it, the policy model predicts the cue's corresponding location in the complete image. Each rollout is evaluated by a rule-based reward that combines format validity and IoU-based localization quality, and the resulting rewards are used to compute group-relative advantages for policy optimization.

We formalize the local visual cue search task as follows. Given a complete image $I$ and a target region $b^*=(x_1,y_1,x_2,y_2)$, we construct a local visual cue by cropping the corresponding region:
\begin{equation}
c = \mathrm{Crop}(I, b^*).
\end{equation}
The resulting crop preserves the target's local appearance while removing most surrounding context. The model is then required to recover the spatial support of this cue in the complete image, which encourages visual matching between the local cue and the full-image context rather than reliance on coarse global semantics.

Formally, the model is provided with the complete image $I$, the visual cue $c$, and an instruction $q$ that asks it to locate the cue in the complete image. It generates a textual response
\begin{equation}
\hat{y} \sim \pi_\theta(\cdot \mid I, c, q),
\end{equation}
from which we parse the predicted bounding box
\begin{equation}
\hat{b} = \mathrm{Parse}(\hat{y}).
\end{equation}
This task differs from conventional text-based grounding in the form of the query. Instead of describing the target with a linguistic expression, the query itself is a local visual cue extracted from the image. The model must therefore compare the appearance of $c$ with the full-image context and recover its spatial support in $I$. This formulation turns grounding into a local visual search problem, where success requires identifying the region in the full image that corresponds to the given cue.

Importantly, the visual cue is used only during training. At inference time, \name{} operates on standard full-image inputs without external cropping, zooming, or multi-round visual search.

\subsection{Reward-Guided Policy Optimization}
\label{sec:reward_optimization}

For the predicted box $\hat{b}$ parsed from the model response, we assign a rule-based reward to measure cue localization quality. Since token-level likelihood does not directly reflect the spatial quality of the predicted region, this reward provides direct supervision on whether the response recovers the cue's spatial support in the complete image. The reward consists of two components. The format reward $r_{\mathrm{format}}$ encourages the model to produce a valid coordinate format, while the localization reward $r_{\mathrm{loc}}$ measures the spatial overlap between the predicted box and the ground-truth box:
\begin{equation}
r_{\mathrm{loc}} =
\begin{cases}
\mathrm{IoU}(\hat{b}, b^*), & \text{if } \hat{b} \text{ is valid}, \\
0, & \text{otherwise}.
\end{cases}
\end{equation}
The final reward is defined as:
\begin{equation}
r(\hat{y}, b^*) = (1-\alpha) r_{\mathrm{loc}} + \alpha r_{\mathrm{format}},
\end{equation}
where $\alpha$ controls the trade-off between localization quality and format validity.

The policy optimization objective is:
\begin{equation}
\max_{\theta}\;
\mathbb{E}_{(I,c,q,b^*) \sim \mathcal{D},\; \hat{y}\sim\pi_{\theta}(\cdot|I,c,q)}
\left[
r(\hat{y}, b^*)
\right],
\end{equation}
where $\mathcal{D}$ denotes the constructed local visual cue search dataset. In practice, we optimize this objective with group-relative policy optimization (GRPO)~\cite{shao2024deepseekmathpushinglimitsmathematical}, using multiple rollouts for each input to compute relative advantages and regularizing the policy against a reference model. By assigning reward according to spatial overlap, the training signal directly encourages the model to localize visual evidence within full-image contexts.

\section{Experiments}
\label{sec:experiments}

\subsection{Implementation Details}

\paragraph{Training Settings.}
We construct the visual-cue search data from COCO train2014 object regions~\cite{lin2014microsoft}. Each training example consists of a complete image, a local crop used as the visual cue, and the corresponding ground-truth bounding box in the original image. The training corpus contains approximately 100K examples, with tiny and small target regions sampled at a 70\%:30\% ratio by default. We adopt Qwen2.5-VL-7B-Instruct~\cite{bai2025qwen25vltechnicalreport} as the primary backbone, and further evaluate \name{} on Qwen3-VL-4B-Thinking~\cite{bai2025qwen3vltechnicalreport} and MiMo-VL-RL-7B~\cite{yue2025mimo} to assess its generality across model families. We perform reinforcement post-training using the EasyR1~\cite{zheng2025easyr1} framework with the GRPO objective and KL-regularized policy optimization. The reward combines format validity with IoU-based localization quality, as described in \S\ref{sec:reward_optimization}. All implementation details and hyperparameters are provided in Appendix B.1.


\paragraph{Evaluation Setup.}
We evaluate \name{} across fine-grained perception, hallucination robustness, general perception, and reasoning benchmarks. Fine-grained perception is assessed on V*Bench~\cite{wu2024v}, HR-Bench-4K/8K~\cite{wang2025divide}, CV-Bench~\cite{tong2024cambrian}, and MME-RealWorld-EN~\cite{zhang2025mme}; hallucination on POPE~\cite{li2023evaluating} and HallusionBench~\cite{guan2024hallusionbench}; general understanding on MMStar~\cite{chen2024we}, RealWorldQA~\cite{ai2024grok}, OCRBench~\cite{liu2024ocrbench}, ScienceQA-IMG~\cite{lu2022learn}, and BabyVision~\cite{chen2026babyvision}; and reasoning on MathVision~\cite{wang2024measuring}, MathVerse~\cite{zhang2024mathverse}, WeMath~\cite{qiao2025we}, and LogicVista~\cite{xiao2024logicvista}. For Qwen2.5-VL-7B, we compare with Vision-R1~\cite{zhan2025vision}, Visual Jigsaw~\cite{wu2025visualjigsawposttrainingimproves}, and PixelReasoner$^\dagger$~\cite{wang2025pixelreasonerincentivizingpixelspace} using the same evaluation pipeline. PixelReasoner$^\dagger$ uses single-pass full-image inference with visual tools disabled. Additional grounding results are provided in Appendix D.3, and full evaluation details in Appendix B.3.

\subsection{Main Results}

\begin{table*}[t]
\centering
{\small
\setlength{\tabcolsep}{1.5pt}
\begin{tabular}{llcccccccccccc}
\toprule
& & \multicolumn{5}{c}{\textbf{Fine-Grained Perception}} & \multicolumn{2}{c}{\textbf{Hallucination}} & \multicolumn{5}{c}{\textbf{General Perception}} \\
\cmidrule(lr){3-7} \cmidrule(lr){8-9} \cmidrule(lr){10-14}
\textbf{Model} & \textbf{Size} & \textbf{V*} & \textbf{HR-4K} & \textbf{HR-8K} & \textbf{CV-B} & \textbf{MME-RW} & \textbf{POPE} & \textbf{HalBench} & \textbf{MMStar} & \textbf{RWQA} & \textbf{OCRBench} & \textbf{SQA$^I$} & \textbf{BabyVision} \\
\midrule
\multicolumn{14}{l}{\textit{Closed-Source Models}} \\
GPT-5.1 & -- & 70.2 & 67.0 & 65.3 & 84.2 & 64.0 & -- & -- & 71.6 & -- & -- & -- & 13.9 \\
Gemini-3-Flash & -- & 86.4 & 87.9 & 85.0 & 89.6 & 74.9 & -- & -- & 83.6 & -- & -- & -- & 34.5 \\
\midrule
Qwen2.5-VL & 7B & 79.6 & 69.9 & 63.8 & 75.6 & 58.8 & 84.9 & 68.0 & 61.5 & 61.6 & 82.0 & 88.7 & 11.9 \\
Vision-R1& 7B & 44.0 & 52.0 & 42.0 & 72.8 & 46.2 & 84.7 & 68.8 & 61.4 & 61.2 & 80.6 & 87.6 & 12.1 \\
Visual Jigsaw & 7B & 83.3 & 71.4 & 66.9 & \textbf{77.6} & 60.8 & 85.3 & 68.5 & 61.1 & 64.8 & 82.3 & 87.2 & \textbf{13.1} \\
PixelReasoner$^\dagger$ & 7B & 81.2 & 70.5 & 64.1 & 76.6 & 61.9 & 86.8 & 67.6 & 63.6 & 63.4 & 84.3 & 89.3 & 10.6 \\
\textbf{\name{}} & 7B & \textbf{87.4} & \textbf{71.6} & \textbf{68.4} & 76.7 & \textbf{62.7} & \textbf{87.6} & \textbf{70.3} & \textbf{63.9} & \textbf{66.4} & \textbf{85.4} & \textbf{89.4} & 12.9 \\
\midrule
Qwen3-VL & 4B & 79.6 & 75.4 & 70.6 & 83.4 & 60.0 & 88.1 & 75.4 & 66.5 & 69.5 & \textbf{78.4} & 92.9 & 9.5 \\
\textbf{\name{}} & 4B & \textbf{82.7} & \textbf{77.0} & \textbf{71.9} & \textbf{84.9} & \textbf{62.0} & \textbf{88.2} & \textbf{75.6} & \textbf{69.7} & \textbf{71.0} & 77.7 & \textbf{93.9} & \textbf{12.1} \\
\midrule
MiMo-VL & 7B & 77.0 & 69.5 & 66.4 & \textbf{80.4} & 54.4 & 86.1 & 67.7 & 66.3 & 66.7 & 81.3 & 93.5 & 9.3 \\
\textbf{\name{}} & 7B & \textbf{79.1} & \textbf{72.0} & \textbf{69.0} & \textbf{80.4} & \textbf{57.1} & \textbf{86.2} & \textbf{72.0} & \textbf{68.9} & \textbf{67.3} & \textbf{82.7} & \textbf{93.7} & \textbf{10.1} \\
\bottomrule
\end{tabular}
}
\caption{Main results across fine-grained perception, hallucination, and general understanding benchmarks. Closed-source models are included for reference, and the Qwen2.5-VL block includes representative 7B post-training baselines. Bold marks the best score within each open-source backbone block. $^\dagger$ denotes single-pass full-image evaluation without visual tools. V*: V*Bench; HR-4K/8K: HR-Bench; CV-B: CV-Bench; MME-RW: MME-RealWorld-EN; HalBench: HallusionBench; RWQA: RealWorldQA; SQA$^{I}$: ScienceQA-IMG.}
\label{tab:main_results}
\end{table*}


\paragraph{Results on perception and hallucination benchmarks.}
Table~\ref{tab:main_results} reports the main results across fine-grained perception, hallucination, and general multimodal understanding benchmarks. On Qwen2.5-VL-7B, \name{} also outperforms representative vision-centric post-training baselines on most evaluated benchmarks. In particular, it reaches 87.4 on V*Bench, 68.4 on HR-Bench-8K, and 62.7 on MME-RealWorld, outperforming the strongest post-training baselines by 4.1, 1.5, and 0.8 points, respectively. Beyond fine-grained perception, \name{} also delivers broad gains over the backbone on POPE, HallusionBench, MMStar, RealWorldQA, OCRBench, ScienceQA-IMG, and BabyVision. The gains further transfer across model families: \name{} improves 11 benchmarks on Qwen3-VL-4B, with only a minor decrease on OCRBench, and improves 11 benchmarks on MiMo-VL-7B while maintaining CV-Bench performance. Together, these results suggest that local visual cue search improves localization-sensitive perception without degrading broad multimodal understanding.

\begin{table}[t]
\centering
{\small
\setlength{\tabcolsep}{1.2pt}
\begin{tabular}{llcccc}
\toprule
\textbf{Model} & \textbf{Size} & \textbf{MathVision} & \textbf{MathVerse} & \textbf{WeMath} & \textbf{LogicVista} \\
\midrule
Qwen2.5-VL & 7B & 23.4 & 46.1 & 64.4 & 43.1 \\
\textbf{+ \name{}} & \textbf{7B} &
\textbf{24.7}$^{\uparrow 1.3}$ &
\textbf{46.6}$^{\uparrow 0.5}$ &
\textbf{65.1}$^{\uparrow 0.7}$ &
\textbf{47.1}$^{\uparrow 4.0}$ \\
\midrule
Qwen3-VL & 4B & 43.4 & 61.4 & 76.3 & 52.5 \\
\textbf{+ \name{}} & \textbf{4B} &
\textbf{44.7}$^{\uparrow 1.3}$ &
\textbf{61.2}$^{\downarrow 0.2}$ &
\textbf{78.8}$^{\uparrow 2.5}$ &
\textbf{56.7}$^{\uparrow 4.2}$ \\
\midrule
MiMo-VL & 7B & 52.3 & 54.2 & 77.2 & 52.5 \\
\textbf{+ \name{}} & \textbf{7B} &
\textbf{54.6}$^{\uparrow 2.3}$ &
\textbf{57.0}$^{\uparrow 2.8}$ &
\textbf{78.3}$^{\uparrow 1.1}$ &
\textbf{55.8}$^{\uparrow 2.3}$ \\
\bottomrule
\end{tabular}
}
\caption{Results on reasoning benchmarks. \name{} preserves mathematical and logical reasoning capabilities across all backbones.}
\label{tab:reasoning}
\end{table}

\paragraph{Results on reasoning benchmarks.}
Table~\ref{tab:reasoning} further evaluates whether \name{} affects mathematical and logical reasoning capabilities. Across MathVision, MathVerse, WeMath, and LogicVista, \name{} largely preserves or improves reasoning performance for all three backbones. For example, \name{} improves LogicVista by 4.0 points on Qwen2.5-VL-7B, 4.2 points on Qwen3-VL-4B, and 2.3 points on MiMo-VL-7B. These results indicate that optimizing local visual cue search does not compromise higher-level reasoning ability; instead, better access to local visual evidence can complement downstream reasoning when visual details are relevant~\cite{tian2025more,sun2026thinking,yang2026look}.

\subsection{Ablations}
\label{sec:exp_ablation}

\paragraph{Effect of training method.}
We study whether the gains of \name{} arise from the visual-cue search data alone or from the optimization objective. Using the same 100K training examples, we compare RL with an SFT baseline using teacher-generated rationales and ground-truth coordinate answers; annotation details are provided in Appendix C.2. As shown in Table~\ref{tab:ablation_sft}, SFT yields only marginal gains over the base model, suggesting that supervised imitation mainly teaches the response format but provides weak pressure for spatial accuracy. In contrast, RL substantially improves all three fine-grained perception benchmarks, increasing V* from 79.6 to 87.4 and HR-8K from 63.8 to 68.4. This indicates that directly optimizing spatial accuracy with an IoU-based reward is crucial for transferring the proxy localization task to downstream fine-grained perception.

\begin{table}[t]
\centering
\small
\setlength{\tabcolsep}{6pt}
\begin{tabular}{lccc}
\toprule
\textbf{Training Method} & \textbf{V*} & \textbf{HR-4K} & \textbf{HR-8K} \\
\midrule
Base & 79.6 & 69.9 & 63.8 \\
SFT & 80.1 & 70.4 & 65.8 \\
\textbf{RL (Ours)} & \textbf{87.4} & \textbf{71.6} & \textbf{68.4} \\
\bottomrule
\end{tabular}
\caption{Ablation on the training method using Qwen2.5-VL-7B and the same 100K visual-cue search data. RL with an IoU-based reward outperforms SFT on fine-grained perception benchmarks.}
\label{tab:ablation_sft}
\end{table}

\begin{table}[t]
\centering
\small
\setlength{\tabcolsep}{6pt}
\begin{tabular}{lcccc}
\toprule
\textbf{Cue Modality} & \textbf{V*} & \textbf{HR-8K} & \textbf{POPE} & \textbf{HalBench} \\
\midrule
Base & 79.6 & 63.8 & 84.9 & 68.0 \\
Text Cue & 83.8 & 67.0 & 85.0 & 69.4 \\
\textbf{Visual Cue (Ours)} & \textbf{87.4} & \textbf{68.4} & \textbf{87.6} & \textbf{70.3} \\
\bottomrule
\end{tabular}
\caption{Ablation on cue modality using Qwen2.5-VL-7B under identical target regions. Text cues use generated referring expressions, while visual cues directly use local crops.}
\label{tab:ablation_modality}
\end{table}

\paragraph{Effect of cue modality.}
We study whether the modality of the cue matters beyond using the same localization-oriented training signal. To ensure a controlled comparison, both variants are trained on identical COCO object regions. For the text-cue baseline, we use Qwen3-VL-235B to generate a referring expression for each target crop, with details provided in Appendix C.3, and train the model to localize the described object in the full image; \name{} instead directly uses the crop as a visual cue. As shown in Table~\ref{tab:ablation_modality}, the text-cue variant improves over the base model, indicating that grounding-style supervision on small target regions is beneficial for fine-grained perception. However, visual cues yield consistently larger gains across all benchmarks, improving V* from 79.6 to 87.4 and HR-8K from 63.8 to 68.4, while also improving POPE and HalBench. This suggests that \name{} benefits not only from localization-oriented supervision, but also from the visual cue itself, which preserves instance-level appearance information that may be difficult to fully capture with generated referring expressions.

\begin{table}[t]
\centering
\small
\setlength{\tabcolsep}{3pt}
\begin{tabular}{lcccc}
\toprule
\textbf{Training Data} & \textbf{Tiny:Small} & \textbf{V*} & \textbf{HR-8K} & \textbf{RefCOCO} \\
\midrule
Base (no training) & -- & 79.6 & 63.8 & 84.4 \\
Large cues ($>$10\%) & -- & 81.2 & 66.4 & \textbf{87.2} \\
\midrule
Mixed cues & 60\%:40\% & 85.9 & 66.8 & 86.2 \\
\textbf{Mixed cues (Ours)} & \textbf{70\%:30\%} & \textbf{87.4} & \textbf{68.4} & 86.5 \\
Mixed cues & 80\%:20\% & 86.9 & 68.1 & 86.3 \\
Mixed cues & 90\%:10\% & \textbf{87.4} & 67.4 & 86.1 \\
Tiny-only cues & 100\%:0\% & 86.9 & 67.0 & 74.4 \\
\bottomrule
\end{tabular}
\caption{Ablation on search difficulty using Qwen2.5-VL-7B. Tiny and small cues denote cue regions with area ratios below 1\% and 1--5\%, respectively; the Tiny:Small column reports their sampling ratio. RefCOCO reports ACC@0.5 averaged across splits. The 70\%:30\% mixture offers the best overall balance.}
\label{tab:ablation_difficulty}
\end{table}

\paragraph{Effect of cue size and search difficulty.}
We study how visual-cue search difficulty affects downstream transfer by varying the size distribution of training cues. Tiny and small cues are defined as regions occupying less than 1\% and 1--5\% of the image area, respectively. As shown in Table~\ref{tab:ablation_difficulty}, large cues improve RefCOCO grounding but yield limited gains on fine-grained perception, suggesting that localizing visually salient regions alone is insufficient for high-resolution evidence retrieval. In contrast, tiny/small cue mixtures substantially improve V* and HR-8K, indicating that harder local search better matches the fine-grained perception challenge. The 70\%:30\% mixture achieves the best HR-8K result and ties for the best V* score while maintaining competitive RefCOCO accuracy. Although the tiny-only variant still performs well on V* and HR-8K, its sharp drop on RefCOCO suggests a scale bias toward small boxes. We therefore use the 70\%:30\% mixture as the default configuration for its best overall trade-off.

\begin{table}[t]
\centering
\small
\setlength{\tabcolsep}{7pt}
\begin{tabular}{lccc}
\toprule
\textbf{Cue Source} & \textbf{V*} & \textbf{HR-4K} & \textbf{HR-8K} \\
\midrule
Base & 79.6 & 69.9 & 63.8 \\
Random Crop & 81.2 & 70.8 & 65.9 \\
\textbf{Object Crop (Ours)} & \textbf{87.4} & \textbf{71.6} & \textbf{68.4} \\
\bottomrule
\end{tabular}
\caption{Ablation on cue source using Qwen2.5-VL-7B. Random-crop cues sampled from arbitrary image regions provide weaker supervision than object-aware cues, highlighting the importance of semantically meaningful local visual cues.}
\label{tab:ablation_random}
\end{table}

\paragraph{Effect of cue source.}
We examine whether the gains of \name{} come from local visual matching alone or from semantically meaningful cues by comparing object-aware cues with random-crop cues sampled from arbitrary image regions rather than object boxes. As shown in Table~\ref{tab:ablation_random}, random crops improve over the base model, indicating that matching local patches to the full image is beneficial. However, they remain substantially weaker than object-aware cues, likely because random crops often contain repeated background textures such as sky, walls, or road surfaces, yielding ambiguous localization signals. In contrast, object-aware cues correspond to coherent visual entities and provide more reliable supervision for fine-grained perception.
\begin{figure}[t]
\centering
\includegraphics[width=\linewidth]{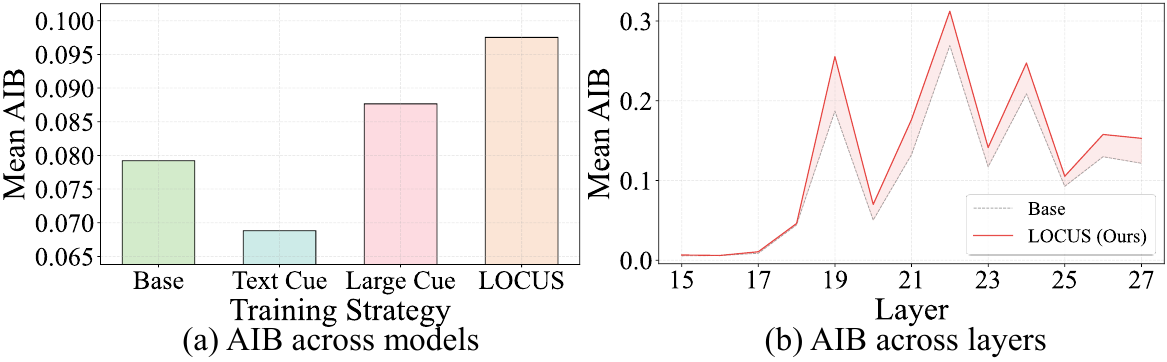}
\caption{Attention-in-Box (AIB) analysis on V*Bench using Qwen2.5-VL-7B variants. (a) compares mean AIB across variants; (b) shows per-layer AIB for Base and \name{}.}
\label{fig:ain_analysis}
\end{figure}

\subsection{Analysis}
\begin{figure*}[t]
\centering
\includegraphics[width=\linewidth]{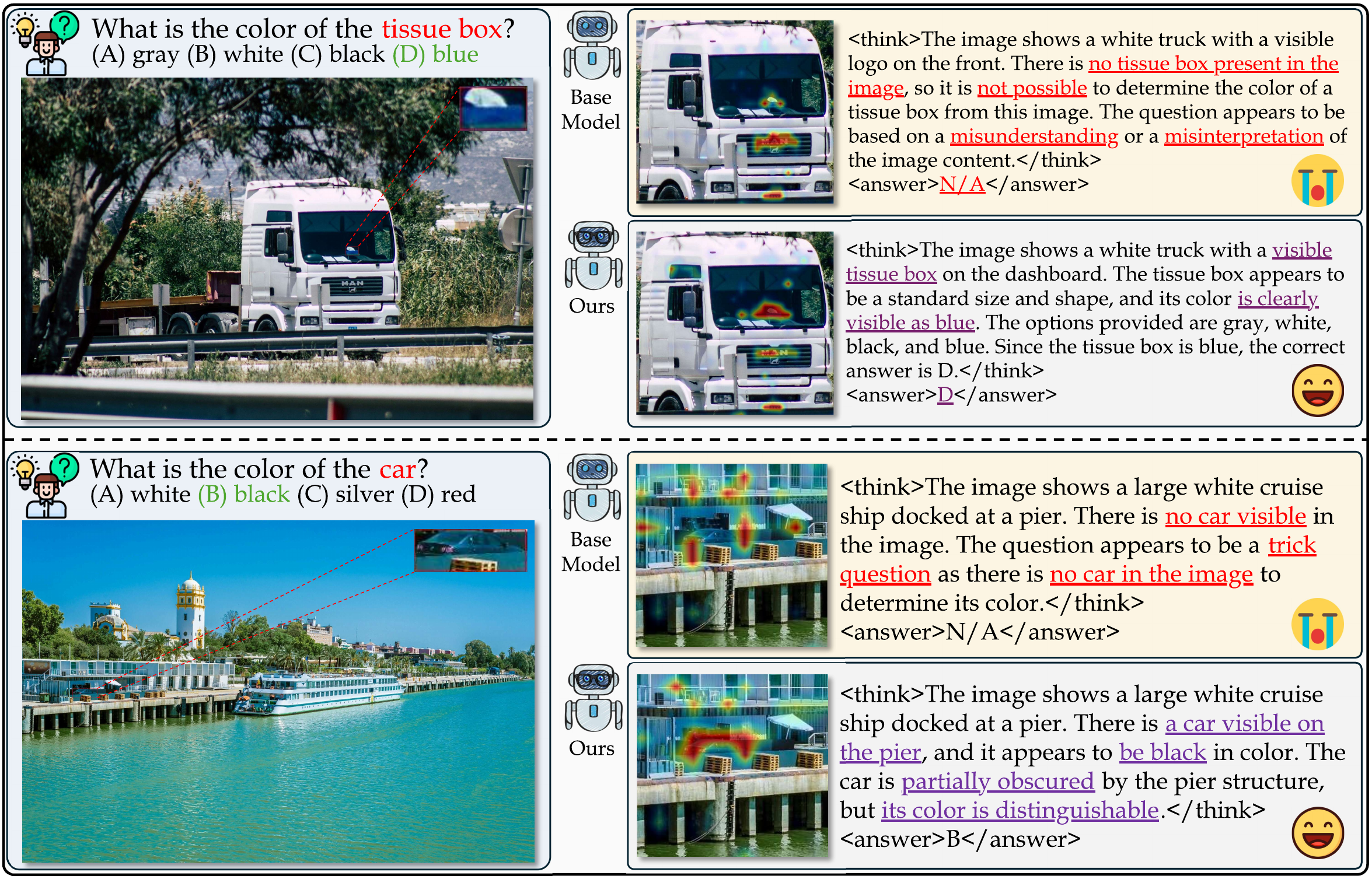}
\caption{Qualitative examples with attention visualizations on V*Bench using Qwen2.5-VL-7B. The base model often fails to select small target evidence and predicts that the queried object is absent, whereas \name{} focuses on the relevant regions and produces the correct answer under standard full-image inference. }
\label{fig:case_study}
\end{figure*}

\paragraph{Attention-in-Box analysis.}
We further analyze whether \name{} changes where the model attends during full-image inference. We compute Attention-in-Box (AIB), defined as the ratio between attention mass within the ground-truth evidence box and total attention mass over all image patches on V*Bench, with implementation details provided in Appendix C.4. As shown in Fig.~\ref{fig:ain_analysis}(a), \name{} achieves the highest mean AIB among all training strategies, outperforming both the base model and the Large Cue variant. Large Cue also improves AIB, but its smaller gain indicates that challenging tiny/small cues provide a stronger signal for learning fine-grained evidence localization. Interestingly, the Text Cue variant improves downstream accuracy but yields lower AIB than the base model, suggesting that text-based cue supervision may help through semantic or language-conditioned alignment rather than attention to the exact evidence region. In contrast, visual-cue search directly trains the model to match a local visual cue against the full-image context, leading to stronger spatial anchoring on the annotated evidence. The layer-wise results in Fig.~\ref{fig:ain_analysis}(b) show comparable AIB in early layers, with a widening gap from around layer 19 onward, where \name{} attends more to the ground-truth box (complete curves in Appendix D.4). This suggests stronger later-stage integration of fine-grained evidence for answer generation, consistent with recent analyses of cross-modal information flow in MLLMs~\cite{zhang2025cross,wu2026vision}, and supports our claim that \name{} improves evidence selection under standard full-image inference.

\paragraph{Qualitative analysis.}
Fig.~\ref{fig:case_study} presents qualitative examples on V*Bench with attention visualizations (details in Appendix C.4). Additional qualitative examples are provided in Appendix D.6. In both cases, the queried object is small and easily overlooked within the full-image context. The base model fails to select the relevant evidence and incorrectly concludes that the target object is absent, producing an invalid answer. In contrast, \name{} places stronger attention on the corresponding local region and correctly identifies the queried object and its color under the same full-image input. These examples illustrate that local visual cue search helps the model recover small but decisive evidence from cluttered scenes without relying on inference-time cropping or zooming.

\section{Conclusion}
\label{sec:conclusion}


We presented \name{}, a training framework that improves fine-grained perception by teaching MLLMs to internalize local visual cue search. Motivated by \emph{visual context rot}, \name{} uses a verifiable proxy task in which the model localizes a cropped visual cue within the original image, optimized with an IoU-based reward. This training-time objective strengthens local evidence selection while preserving the standard image-question inference interface. Experiments across perception, hallucination, general understanding, and reasoning benchmarks show consistent gains on localization-sensitive tasks without degrading broad capabilities. Attention analyses further indicate stronger focus on ground-truth evidence regions, suggesting that cue localization training translates into more reliable evidence use at inference time. These results highlight internal local evidence search as a promising direction for building MLLMs that are more robust to dense, high-resolution visual contexts.

\bibliography{references}

\appendix

\section{Use of Large Language Models}
We used large language models only as writing assistants during manuscript preparation. Specifically, they were used for language polishing, grammar correction, and improving the clarity and readability of the text. They were not used to generate research ideas, design the method or experiments, conduct analyses, or draw scientific conclusions. The authors carefully reviewed all model-assisted edits and retained full responsibility for the final content of the paper.

\section{Implementation Details}

\subsection{Training Configuration}
\label{app:train_config}

\begin{table}[t]
\centering
\small
\begin{tabular}{ll}
\toprule
\textbf{Statistic} & \textbf{Value} \\
\midrule
Training samples & 99,500 \\
Validation samples & 500 \\
Source & COCO train2014 \\
Tiny cue ratio (area $<$ 1\%) & 70\% \\
Small cue ratio (1--5\%) & 30\% \\
Min crop size & 16 px \\
Padding range (tiny cues) & 0--10\% \\
Scale range (tiny cues) & 0.8--2.0$\times$ \\
\bottomrule
\end{tabular}
\caption{Statistics of the default local visual cue search training corpus.}
\label{tab:data_stats}
\end{table}

\begin{table}[t]
\centering
\small
\setlength{\tabcolsep}{5pt}
\begin{tabular}{lll}
\toprule
\textbf{Component} & \textbf{Parameter} & \textbf{Value} \\
\midrule
\multirow{4}{*}{\textbf{Algorithm}}
& RL algorithm & GRPO \\
& KL coefficient & $1.0 \times 10^{-2}$ \\
& Reward & IoU + format reward \\
& Format weight & 0.1 \\
\midrule
\multirow{6}{*}{\textbf{Optimization}}
& Learning rate & $1.0 \times 10^{-6}$ \\
& Weight decay & $1.0 \times 10^{-2}$ \\
& Optimizer & AdamW \\
& Gradient clipping & 1.0 \\
& Warmup ratio & 0.05 \\
& Global batch size & 128 \\
\midrule
\multirow{4}{*}{\textbf{Rollout}}
& Rollouts per prompt & 8 \\
& Temperature & 1.0 \\
& Top-p & 1.0 \\
& Validation override & $T{=}0.6$, $p{=}0.95$, $n{=}1$ \\
\bottomrule
\end{tabular}
\caption{Reinforcement post-training configuration for the primary Qwen2.5-VL-7B \name{} experiment.}
\label{tab:rl_config}
\end{table}
Unless otherwise specified, the implementation details below describe our primary Qwen2.5-VL-7B~\cite{bai2025qwen25vltechnicalreport} experiments, which use the full 100K local visual cue search corpus. Additional backbone experiments follow the same data construction and training objective, but may use model-specific training lengths.

We construct the local visual cue search corpus from COCO train2014 object annotations~\cite{lin2014microsoft}. Each example contains a complete image, a localized visual cue cropped from the same image, and the corresponding ground-truth bounding box in the original image. The default training split contains 99,500 training examples and 500 validation examples. We sample tiny and small cue regions at a 70\%:30\% ratio, where tiny regions occupy less than 1\% of the image area and small regions occupy 1--5\%. For Qwen2.5-VL-7B~\cite{bai2025qwen25vltechnicalreport}, all target boxes are represented using absolute pixel coordinates. The resulting corpus statistics are summarized in Table~\ref{tab:data_stats}.

We perform reinforcement post-training with GRPO using the EasyR1 framework~\cite{zheng2025easyr1,sheng2024hybridflow}. The reward is computed from the model response by parsing the box inside the \texttt{<answer>} tag. The localization reward is the IoU between the predicted box and the ground-truth box, and the final reward combines this IoU reward with a format reward that checks whether the response follows the required \texttt{<think>} and \texttt{<answer>} structure. The format reward weight is 0.1. The main training hyperparameters are listed in Table~\ref{tab:rl_config}. 

All \name{} post-training runs were conducted on a single node equipped with eight NVIDIA A100-80G GPUs. Table~\ref{tab:training_cost} reports the wall-clock time and total GPU-hours for one training run of each backbone. GPU-hours are calculated as the number of GPUs multiplied by wall-clock time. These costs cover reinforcement post-training and exclude benchmark evaluation.

\begin{table}[t]
\centering
\small
\setlength{\tabcolsep}{5pt}
\begin{tabular}{lccr}
\toprule
\textbf{Backbone} & \textbf{GPUs} & \textbf{Time (h)} & \textbf{GPU-hours} \\
\midrule
Qwen2.5-VL-7B & 8 & 31.3 & 250.4 \\
MiMo-VL-7B   & 8 & 33.0 & 264.0 \\
Qwen3-VL-4B  & 8 & 40.0 & 320.0 \\
\bottomrule
\end{tabular}
\caption{Training cost of \name{} for each backbone.}
\label{tab:training_cost}
\end{table}

\subsection{Prompt Template}
The training prompt provides the full image and the localized visual cue as two image inputs, and asks the model to recover the cue location in the full image. The prompt template used for the primary Qwen2.5-VL-7B training is shown below.

\begin{tcolorbox}[
  colback=white,
  colframe=black,
  colbacktitle=white,
  coltitle=black,
  fonttitle=\bfseries,
  title=Prompt for Local Visual Cue Search,
  boxrule=0.7pt,
  arc=2pt,
  left=5pt,
  right=5pt,
  top=4pt,
  bottom=4pt
]
\texttt{<image>}\\
\texttt{<image>}\\
Image 1 is a full scene image. Image 2 is a cropped region from Image 1. Please find where Image 2 is located in Image 1, and output the bounding box as \([x_1, y_1, x_2, y_2]\) in pixel coordinates.

A conversation between User and Assistant. The user asks a question, and the Assistant solves it. The assistant first thinks about the reasoning process in the mind and then provides the user with the answer. The reasoning process and answer are enclosed within \texttt{<think>} \texttt{</think>} and \texttt{<answer>} \texttt{</answer>} tags, respectively, i.e., \texttt{<think>} reasoning process here \texttt{</think>}\texttt{<answer>} answer here \texttt{</answer>}.
\end{tcolorbox}

\subsection{Evaluation Configuration}
\label{app:eval_config}

We evaluate all models with the same benchmark adapters and vLLM-based inference pipeline. Unless otherwise specified, decoding uses greedy generation with temperature 0 and a maximum generation length of 8192 tokens. For checkpoints trained with the reasoning format, we append the same \texttt{<think>}/\texttt{<answer>} instruction to evaluation prompts; for models with native or model-specific thinking formats, we use their corresponding inference templates.

Our evaluation covers fine-grained perception benchmarks, including V*Bench~\cite{wu2024v}, HR-Bench-4K/8K~\cite{wang2025divide}, CV-Bench~\cite{tong2024cambrian}, and MME-RealWorld-EN~\cite{zhang2025mme}; visual grounding benchmarks, including RefCOCO, RefCOCO+~\cite{yu2016modeling,kazemzadeh-etal-2014-referitgame}, and RefCOCOg~\cite{mao2016generation}; hallucination benchmarks, including POPE~\cite{li2023evaluating} and HallusionBench~\cite{guan2024hallusionbench}; general perception benchmarks, including MMStar~\cite{chen2024we}, RealWorldQA~\cite{ai2024grok}, OCRBench~\cite{liu2024ocrbench}, ScienceQA-IMG (SQA$^I$)~\cite{lu2022learn}, and BabyVision~\cite{chen2026babyvision}; and reasoning benchmarks, including MathVision~\cite{wang2024measuring}, MathVerse~\cite{zhang2024mathverse}, WeMath~\cite{qiao2025we}, and LogicVista~\cite{xiao2024logicvista}. For multiple-choice benchmarks, each adapter extracts the final option letter from the generated response. For open-ended perception benchmarks, we apply the benchmark-specific normalization and matching rules implemented in the corresponding adapter; RealWorldQA is handled according to its multiple-choice or open-ended format. For grounding benchmarks, we parse the generated bounding box and report ACC@0.5. Raw model outputs are saved without truncation before answer parsing, which is important for long reasoning outputs. For the Qwen2.5-VL-7B baseline comparison, all models are evaluated using the same benchmark adapters and answer-parsing protocols. PixelReasoner$^\dagger$ is evaluated under a restricted single-round full-image setting: each example contains only the original image and question, and the model generates one response without inference-time cropping, zooming, visual-tool execution, or additional visual observations. This setting matches the standard image-question inference interface of \name{} and evaluates the capability encoded in the checkpoint rather than the additional benefits of tool-augmented inference.

\section{Additional Experimental Details}
\subsection{Training Data Examples}

\begin{figure*}[t]
\centering
\setlength{\tabcolsep}{2pt}
\begin{tabular}{cc}
\includegraphics[width=0.49\textwidth]{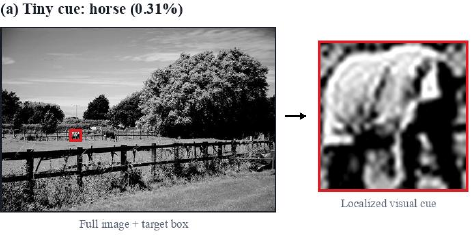} &
\includegraphics[width=0.49\textwidth]{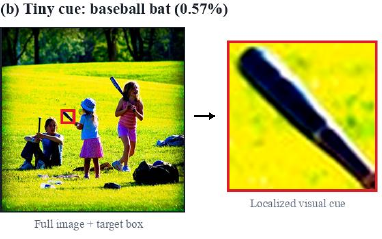} \\
\includegraphics[width=0.49\textwidth]{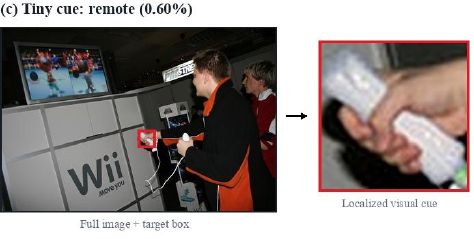} &
\includegraphics[width=0.49\textwidth]{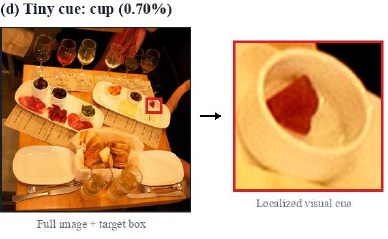} \\
\includegraphics[width=0.49\textwidth]{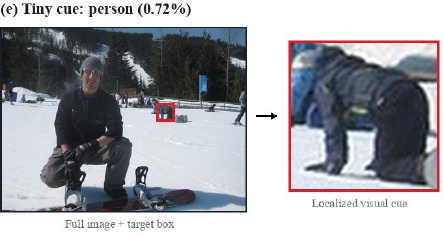} &
\includegraphics[width=0.49\textwidth]{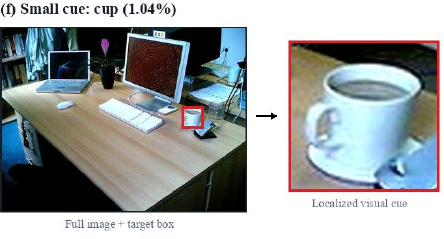} \\
\end{tabular}
\caption{
Examples of local visual cue search training data. Each sample contains a full image with the target box and a localized visual cue cropped from the same image. The task asks the model to recover the cue's spatial support in the full image.
}
\label{fig:training_data_examples}
\end{figure*}

Fig.~\ref{fig:training_data_examples} shows representative examples from our local visual cue search training data. Each sample consists of a full image, a localized visual cue cropped from the same image, and the ground-truth target box of the cue in the full image. The examples cover both tiny and small cue regions, illustrating that the model must recover visually subtle local evidence from cluttered full-image context. The visual cue is used only during training to construct a verifiable localization objective; inference uses the standard image-question input without any crop.

\subsection{SFT Baseline Construction}
\label{app:sft_baseline}

For the SFT baseline in Table~4 of the main paper, we use the same 100K local visual cue search examples as \name{}. To construct reasoning-format supervision, we use Qwen2.5-VL-72B-Instruct~\cite{bai2025qwen25vltechnicalreport} as a teacher model. During annotation, the teacher is given the full image, the crop image, and the ground-truth bounding box, and is asked to generate a brief rationale explaining how the crop can be localized in the full image. We then extract the generated rationale and pair it with the ground-truth bounding box as the final answer, yielding responses in the same \texttt{<think>} and \texttt{<answer>} format as the RL training output.

\begin{tcolorbox}[
  colback=white,
  colframe=black,
  colbacktitle=white,
  coltitle=black,
  fonttitle=\bfseries,
  title=Prompt for SFT Rationale Annotation,
  boxrule=0.7pt,
  arc=2pt,
  left=5pt,
  right=5pt,
  top=4pt,
  bottom=4pt
]
\texttt{<image>}\\
\texttt{<image>}\\
Image 1 is a full scene image. Image 2 is a cropped region from Image 1.\\
The cropped region is located at \texttt{\{bbox\}} in Image 1 (pixel coordinates [x1, y1, x2, y2]).

Please write a brief reasoning process (2--3 sentences) explaining how you would identify where Image 2 is located in Image 1 based on the visual cues in the crop. Then provide the bounding box.

Output format:\\
\texttt{<think>}your reasoning\texttt{</think>}\\
\texttt{<answer>}\texttt{\{bbox\}}\texttt{</answer>}
\end{tcolorbox}

Importantly, the ground-truth box is used only for teacher-side rationale annotation. During SFT training, the student receives the same input as in RL training, namely the full image, the crop image, and the instruction to locate the crop in the full image; the ground-truth box is not included in the prompt. This baseline tests whether supervised imitation of teacher-generated rationales and ground-truth coordinate answers is sufficient, compared with directly optimizing localization quality through the IoU-based RL reward.

\subsection{Text Cue Baseline Construction}
\label{app:text_cue}

The text-cue baseline in Table~5 is designed to isolate the effect of cue modality while keeping the target regions and optimization setup unchanged. We start from the same 100K tiny/small training examples used by \name{}, each containing a full image, a crop corresponding to the target region, and the ground-truth bounding box in pixel coordinates. For each example, we use Qwen3-VL-235B-A22B-Instruct~\cite{bai2025qwen3vltechnicalreport} to generate a short referring expression conditioned on both the full image and the target crop. The annotation prompt asks the model to produce a single concise expression under 25 words that uniquely identifies the cropped object in the full image, focusing on category, appearance, size, and spatial relations to nearby objects.

\begin{tcolorbox}[
  colback=white,
  colframe=black,
  colbacktitle=white,
  coltitle=black,
  fonttitle=\bfseries,
  title=Prompt for Text Cue Annotation,
  boxrule=0.7pt,
  arc=2pt,
  left=5pt,
  right=5pt,
  top=4pt,
  bottom=4pt
]
You are given a full image and a cropped region from it. The cropped region highlights a specific object in the scene.

Write a short, precise referring expression (1 sentence, under 25 words) that uniquely identifies this object in the full image. Focus on its category, appearance, size, and spatial position relative to other objects.

Output only the referring expression, nothing else.
\end{tcolorbox}

After generation, each sample is converted into a text-grounding training instance by replacing the visual crop with the generated referring expression. The model receives only the full image and a text query of the form: ``Please find the object described by the following text in the image,'' followed by the referring expression, and is trained to output the same ground-truth box as \([x_1, y_1, x_2, y_2]\) in pixel coordinates. The response format follows the same reasoning template as \name{}, with intermediate reasoning enclosed by \texttt{<think>} and the final box enclosed by \texttt{<answer>}.

This construction ensures a controlled comparison between text and visual cues: both variants use identical images, target boxes, data split, backbone, GRPO training configuration, and IoU-based localization reward. The only difference is the query modality. The text-cue baseline uses a generated linguistic description as the localization query, whereas \name{} uses the local visual crop itself.

\subsection{Attention-in-Box Analysis}
\label{app:aib}

We use Attention-in-Box (AIB) to quantify whether a model allocates more attention to the ground-truth evidence region during ordinary full-image inference. The analysis is conducted on V*Bench samples with manually annotated object bounding boxes. For each model and each sample, we run the same image-question prompt as in evaluation and extract the attention from the position that predicts the first answer token to all image-patch tokens. We use HuggingFace inference with \texttt{output\_attentions=True}; the language-model attention uses the eager implementation to expose attention weights, while the vision encoder uses memory-efficient attention. Image preprocessing follows the evaluation pipeline, including the same chat template, \texttt{qwen\_vl\_utils} image processing, and processor pixel limits.

For each layer, we average attention weights over heads and keep only the entries corresponding to image-patch tokens. These values are reshaped to the spatial image-token grid derived from \texttt{image\_grid\_thw} after the model's spatial merge. Let \(A_l \in \mathbb{R}^{H \times W}\) denote the resulting attention map at layer \(l\). We project the ground-truth bounding box onto the same grid and construct a binary mask \(M \in \{0,1\}^{H \times W}\), where a grid cell is included if it intersects any annotated box. The layer-wise AIB is defined as
\begin{equation}
\mathrm{AIB}_l = \frac{\sum_{i,j} A_l(i,j) M(i,j)}{\sum_{i,j} A_l(i,j)}.
\end{equation}
The mean AIB reported in Fig.~4(a) of the main paper is computed from the layer-averaged attention map, while Fig.~4(b) reports \(\mathrm{AIB}_l\) across layers. We also compute Peak-in-Box, which checks whether the maximum-attention image token lies inside the ground-truth box, but use AIB as the primary metric because it measures total attention mass assigned to the evidence region.

For qualitative visualization, we upsample the layer-averaged attention map to the original image resolution with bilinear interpolation and overlay it on the image as a heatmap. To reduce visual clutter, low-attention pixels below a percentile threshold are rendered transparent, and the remaining values are shown with a bounded opacity. For side-by-side comparisons, Base and \name{} heatmaps are normalized with shared peak-ratio scaling so that color intensity remains comparable across the two models. Ground-truth boxes are drawn on the original image to indicate the annotated evidence region.

\section{Additional Results}

\subsection{Grounding--VQA Correlation on Qwen3-VL}
\label{add:grounding_VQA}
\begin{figure}[t]
  \centering
  \includegraphics[width=\linewidth]{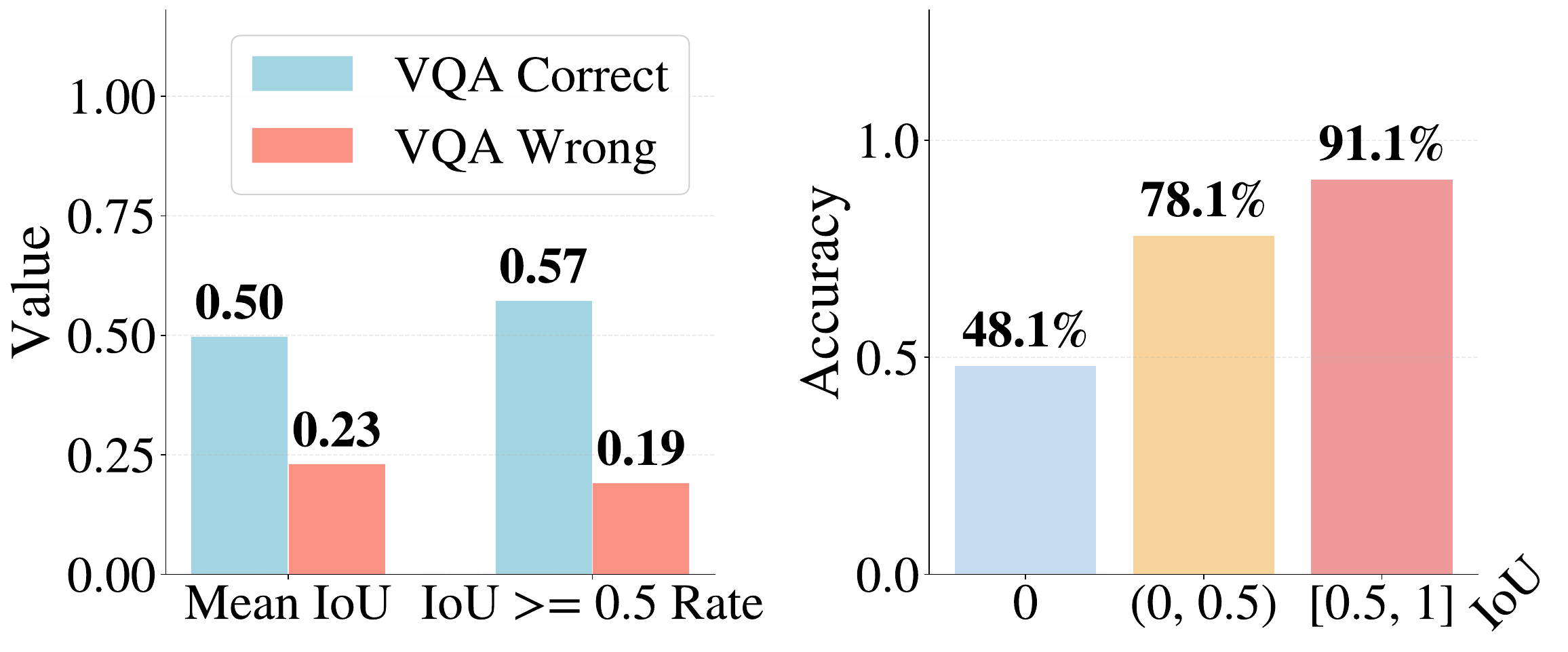}
 \caption{Additional grounding--VQA correlation analysis for Qwen3-VL-4B-Thinking on the V*Bench direct-attribute subset. (a) Mean IoU and grounding success rate (IoU $\ge$ 0.5) for VQA-correct vs. VQA-wrong samples. (b) VQA accuracy across grounding IoU levels.}
  \label{fig:motivation_qwen3}
\end{figure}

To examine whether the correlation between localization quality and fine-grained VQA correctness also holds beyond Qwen2.5-VL, we repeat the analysis in Fig.~2 of the main paper using Qwen3-VL-4B-Thinking on the V*Bench direct-attribute subset. As shown in Fig.~\ref{fig:motivation_qwen3}, correctly answered samples exhibit higher grounding IoU and grounding success rate than incorrectly answered samples, and VQA accuracy generally increases with grounding IoU. This provides additional evidence that reliable localization of task-relevant regions is closely associated with fine-grained perception.

\subsection{Validation Performance on Local Visual Cue Search}

\begin{table}[t]
\centering
\small
\setlength{\tabcolsep}{2.5pt}
\begin{tabular}{lcccc}
\toprule
\textbf{Model} & \textbf{Size} & \textbf{Mean IoU} & \textbf{ACC@0.5} & \textbf{ACC@0.75} \\
\midrule
Qwen2.5-VL & 7B & 16.4 & 16.4 & 9.4 \\
\textbf{+ \name{}} & \textbf{7B} & \textbf{43.0}$^{\uparrow26.6}$ & \textbf{48.8}$^{\uparrow32.4}$ & \textbf{25.0}$^{\uparrow15.6}$ \\
\midrule
Qwen3-VL & 4B & 21.7 & 19.6 & 11.2 \\
\textbf{+ \name{}} & \textbf{4B} & \textbf{43.2}$^{\uparrow21.5}$ & \textbf{48.4}$^{\uparrow28.8}$ & \textbf{31.0}$^{\uparrow19.8}$ \\
\bottomrule
\end{tabular}
\caption{Local visual cue search performance on the held-out validation set (500 samples, 70\% tiny / 30\% small cues). \name{} substantially improves localization accuracy across both backbones, confirming that the model learns precise visual matching through RL training.}
\label{tab:rg_val}
\end{table}

We further evaluate whether \name{} directly improves the proxy task it is trained on, namely local visual cue search. Table~\ref{tab:rg_val} reports localization performance on the held-out validation split of our cue-search data, consisting of 500 samples with the same 70\% tiny and 30\% small cue distribution as the main training setting. The base models show limited zero-shot ability to match a localized visual cue back to its spatial support in the full image, achieving only 16.4 and 21.7 mean IoU for Qwen2.5-VL-7B and Qwen3-VL-4B, respectively. This indicates that local visual cue search is a non-trivial capability even when the target crop is explicitly provided.

After \name{} training, both backbones obtain large gains across all localization metrics. Qwen2.5-VL improves from 16.4 to 43.0 in mean IoU and from 16.4 to 48.8 in ACC@0.5, while Qwen3-VL improves from 21.7 to 43.2 in mean IoU and from 19.6 to 48.4 in ACC@0.5. These results confirm that the RL objective effectively teaches the model to perform precise local visual matching, providing direct evidence that the downstream gains are grounded in improved cue localization ability.

\subsection{Additional Visual Grounding Results}

We additionally report visual grounding results on RefCOCO, RefCOCO+~\cite{yu2016modeling,kazemzadeh-etal-2014-referitgame}, and RefCOCOg~\cite{mao2016generation} in Table~\ref{tab:grounding_appendix}. These results are provided as a supplementary evaluation of localization ability under standard text-based referring expressions. On Qwen2.5-VL-7B, \name{} improves the average ACC@0.5 from 81.4 to 82.7, indicating that local visual cue search does not harm general grounding ability while improving fine-grained perception.

\begin{table}[t]
\centering
\small
\setlength{\tabcolsep}{3pt}
\begin{tabular}{lcccc}
\toprule
\textbf{Model} & \textbf{RefCOCO} & \textbf{RefCOCO+} & \textbf{RefCOCOg} & \textbf{Avg.} \\
\midrule
Qwen2.5-VL & 84.4 & 78.4 & 81.1 & 81.4 \\
\textbf{+ \name{}} & \textbf{86.5}$^{\uparrow2.1}$ & \textbf{78.8}$^{\uparrow0.4}$ & \textbf{83.0}$^{\uparrow1.9}$ & \textbf{82.7}$^{\uparrow1.3}$ \\
\bottomrule
\end{tabular}
\caption{Additional visual grounding results for Qwen2.5-VL-7B on RefCOCO, RefCOCO+, and RefCOCOg. We report ACC@0.5 averaged across splits.}
\label{tab:grounding_appendix}
\end{table}

\subsection{Complete Layer-wise AIB Curves}
\label{app:aib_full_layer}

\begin{figure}[t]
  \centering
  \includegraphics[width=\linewidth]{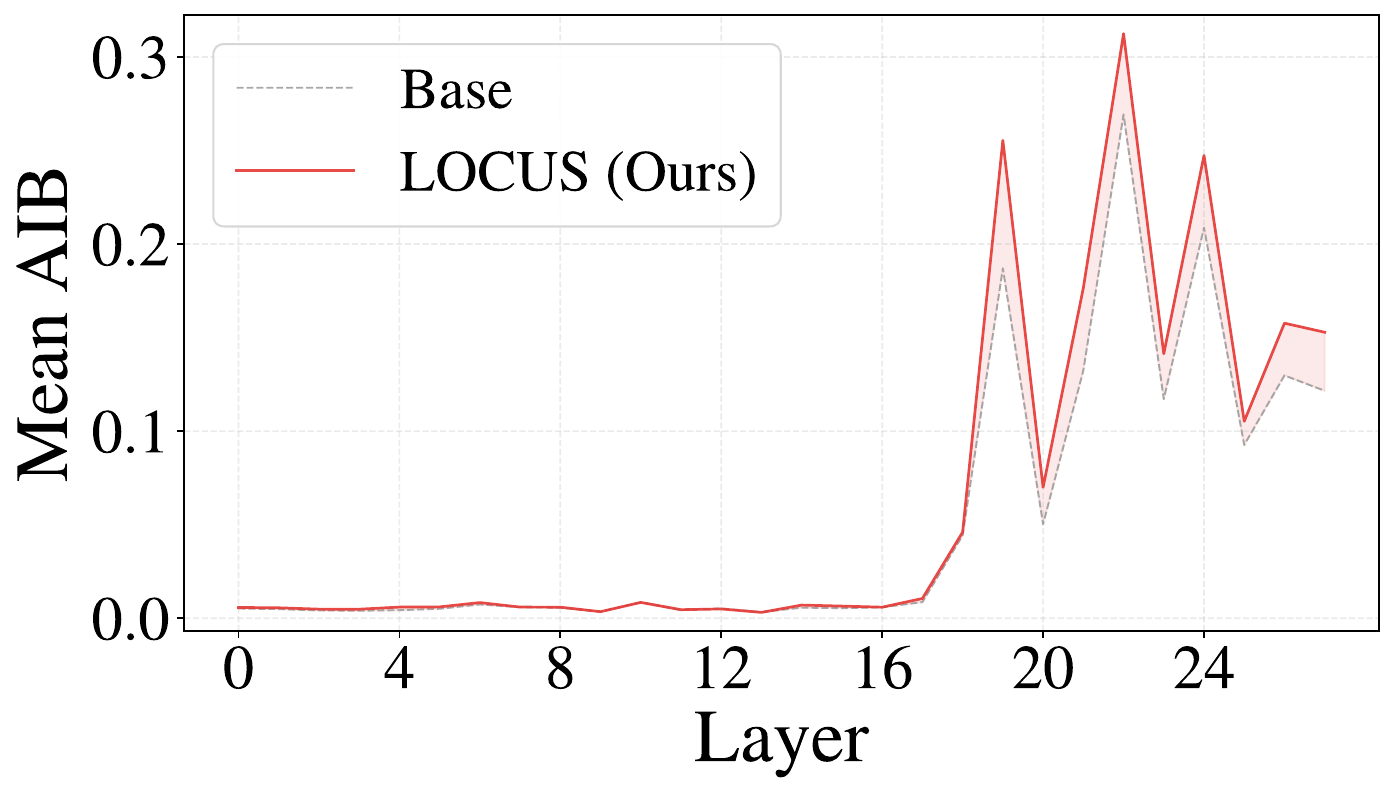}
\caption{
Complete layer-wise Attention-in-Box (AIB) curves on V*Bench using Qwen2.5-VL-7B. The AIB gap between Base and \name{} becomes more pronounced in later layers.
}
  \label{fig:aib_full_layer}
\end{figure}

Fig.~\ref{fig:aib_full_layer} provides the complete layer-wise Attention-in-Box (AIB) curves on V*Bench. The two models exhibit similar AIB values in early layers, while the gap becomes substantially larger from around layer 19 onward. This pattern suggests that the integration of fine-grained visual evidence for answer generation is more pronounced in later layers, where \name{} assigns more attention mass to the ground-truth evidence region than the base model. 

\subsection{Training Dynamics}

\begin{figure}[t]
\centering
\includegraphics[width=\linewidth]{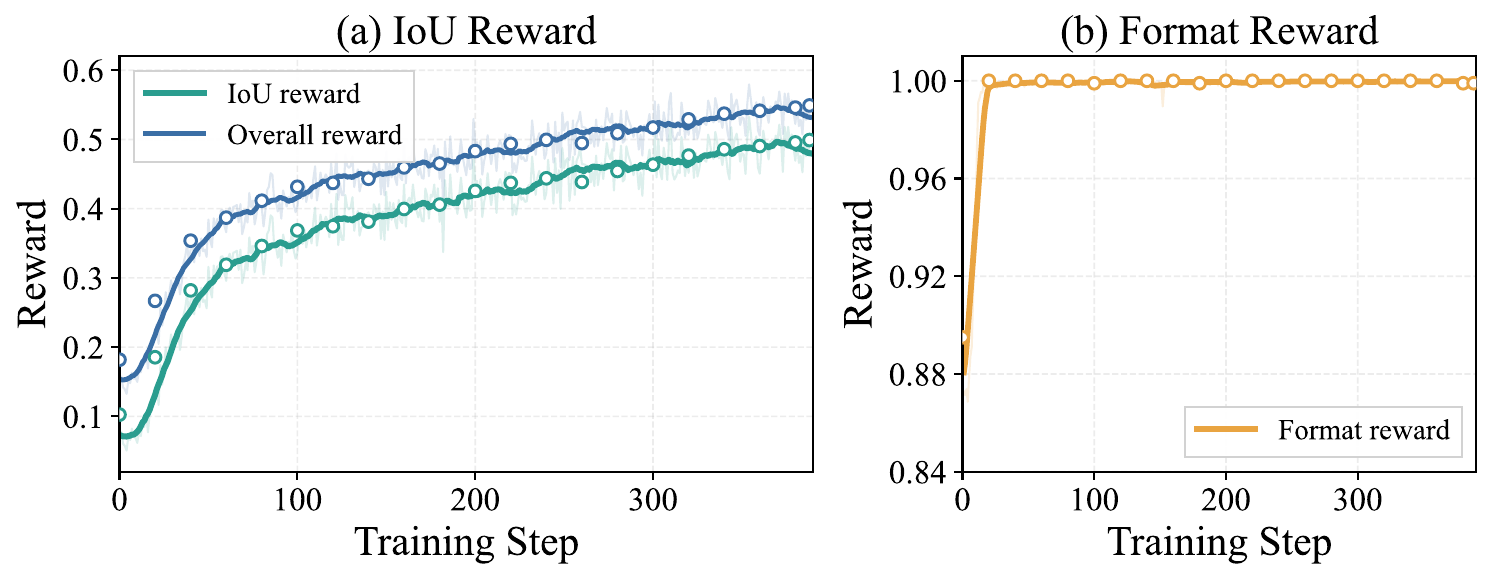}
\caption{Training dynamics of the primary Qwen2.5-VL-7B \name{} experiment. Faint lines show raw training rewards, solid lines show moving averages, and hollow markers denote validation rewards.}
\label{fig:training_dynamics}
\end{figure}

We show the training dynamics of the primary Qwen2.5-VL-7B \name{} experiment in Fig.~\ref{fig:training_dynamics}. The IoU-based localization reward steadily improves during GRPO training, while the format reward quickly converges, indicating stable optimization of the local visual cue search objective.

\begin{figure*}[t]
  \centering
  \includegraphics[width=\textwidth]{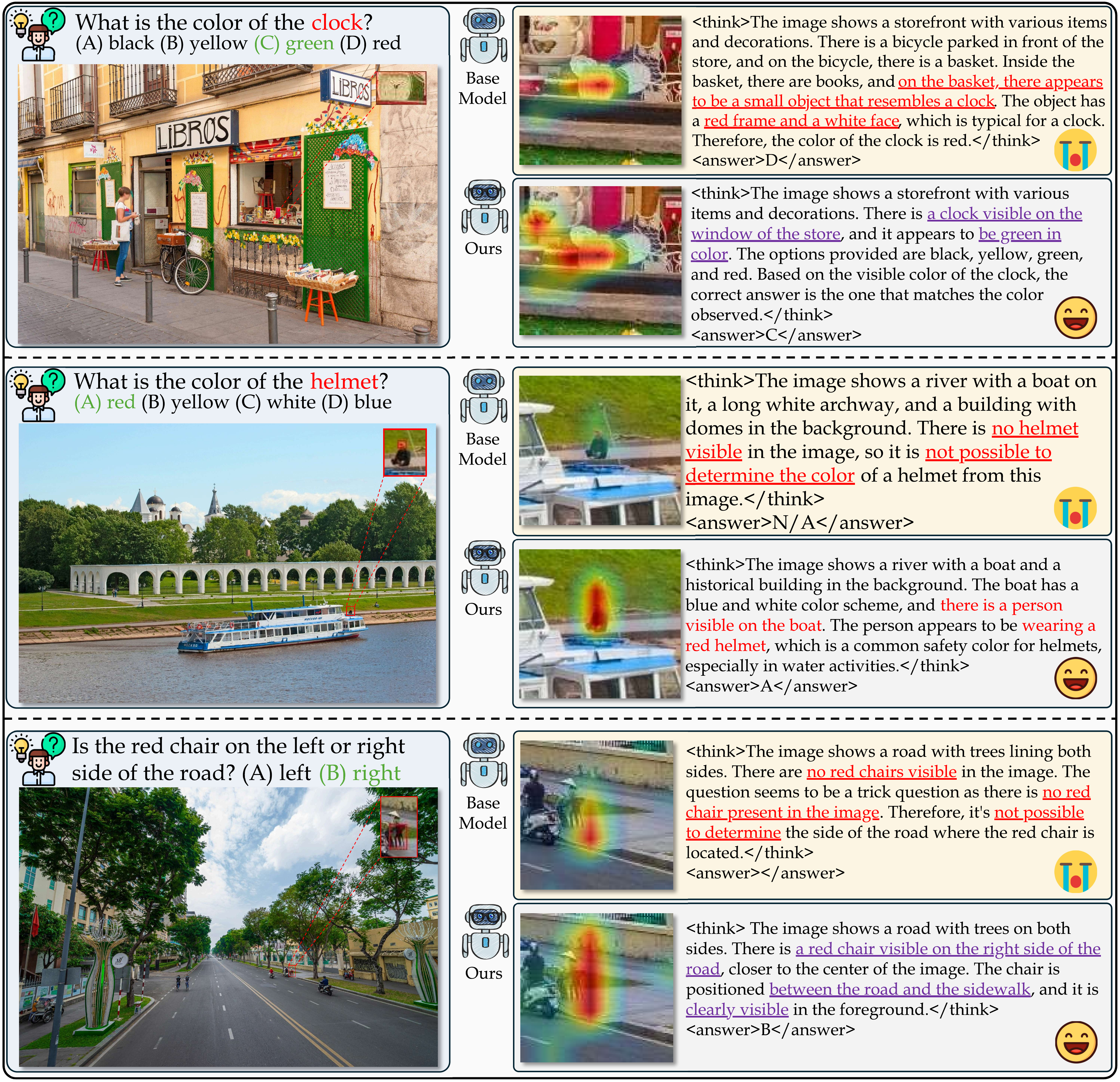}
  \caption{
  Additional qualitative examples on V*Bench using Qwen2.5-VL-7B. The base model often overlooks small target evidence or incorrectly concludes that the queried object is absent, whereas \name{} focuses on the relevant local region and produces the correct answer under the same full-image input.
  }
  \label{fig:additional_case_study}
\end{figure*}

\subsection{Additional Qualitative Examples}

Fig.~\ref{fig:additional_case_study} provides additional qualitative examples on V*Bench. Across these cases, the queried evidence is small and easily overlooked within the full-image context. The base model often fails to locate the target object and either predicts an incorrect attribute or concludes that the object is absent. In contrast, \name{} attends more strongly to the relevant local region and recovers the correct answer under the same full-image input. These examples further illustrate that local visual cue search improves fine-grained perception by strengthening evidence retrieval without requiring inference-time cropping or zooming.

\subsection{Failure Cases}

\begin{figure*}[t]
  \centering
  \includegraphics[width=\textwidth]{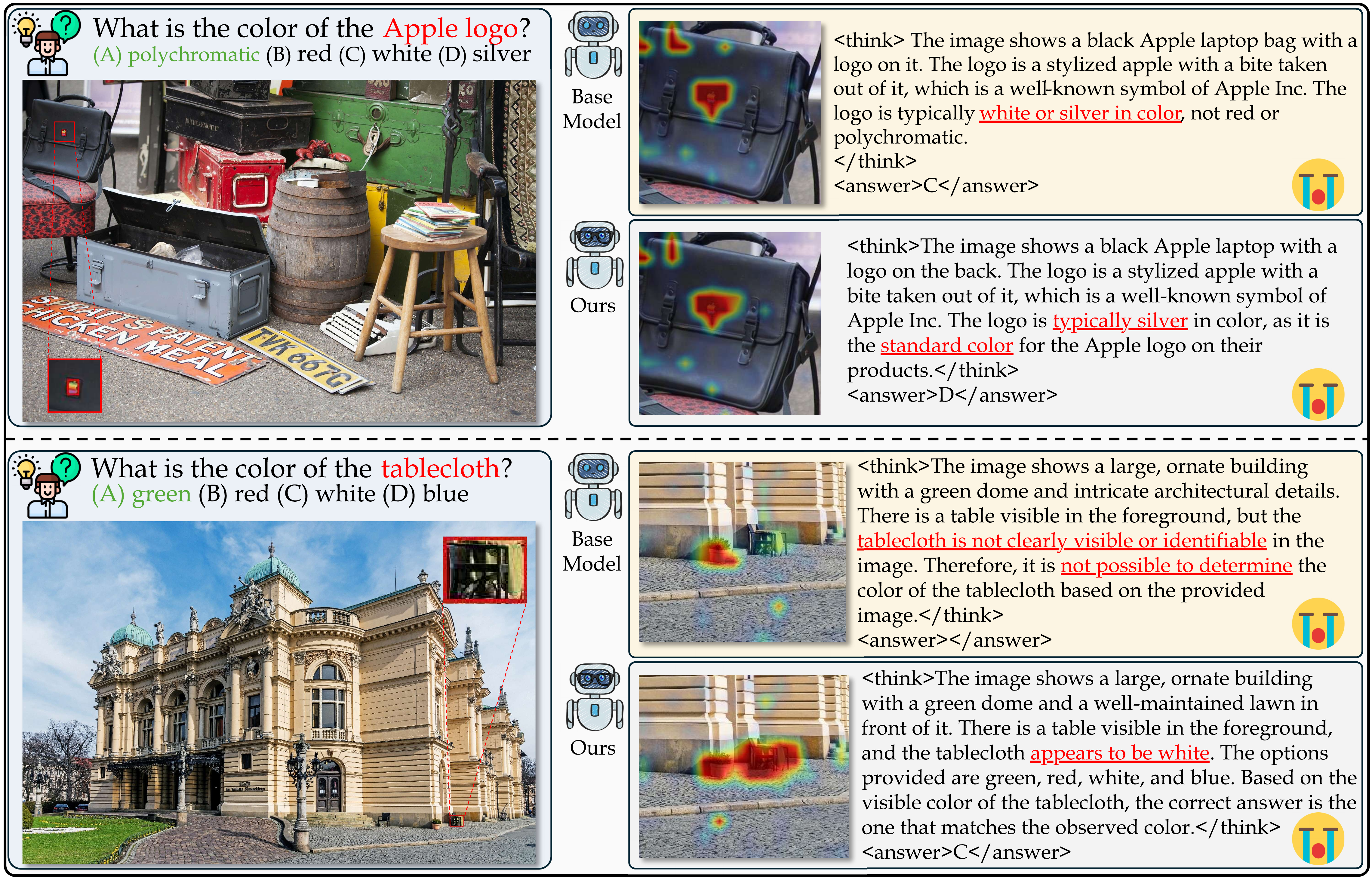}
  \caption{
  Representative failure cases on V*Bench using Qwen2.5-VL-7B. Although \name{} can attend to the relevant local region, it may still fail when the retrieved evidence requires subtle attribute interpretation or semantic disambiguation. These examples indicate that fine-grained perception involves both local evidence retrieval and accurate visual understanding.
  }
  \label{fig:failure_cases}
\end{figure*}

Although \name{} improves the model's ability to retrieve local evidence, fine-grained perception also requires correctly interpreting the retrieved visual content. Fig.~\ref{fig:failure_cases} shows representative failure cases where both the base model and \name{} produce incorrect answers. In these examples, the models often attend to or describe the relevant region, but still fail to infer the correct fine-grained attribute. For instance, in the Apple-logo example, both models focus on the logo area but interpret its color as a single dominant color rather than recognizing it as polychromatic. Similarly, in the tablecloth example, \name{} attends to the local region around the tablecloth, yet misidentifies its color under cluttered background and low-resolution visual evidence.

These cases suggest that local visual cue search mainly addresses the evidence retrieval bottleneck, but does not fully solve all fine-grained perception failures. Accurate answers may still depend on attribute-level recognition, semantic disambiguation, and robust interpretation of small or visually ambiguous regions. We therefore view \name{} as a complementary mechanism that improves access to decisive evidence, while stronger visual understanding remains necessary for resolving cases where the evidence itself is difficult to interpret.

\end{document}